\documentclass[10pt,twocolumn,letterpaper]{article}

\usepackage{cvpr}              

\usepackage[dvipsnames]{xcolor}

\newcommand\lft{\mathopen{}\left}
\newcommand\rgt{\aftergroup\mathclose\aftergroup{\aftergroup}\right}

\definecolor{tabfirst}{rgb}{1, 0.7, 0.7}
\definecolor{tabsecond}{rgb}{1, 0.85, 0.7}
\definecolor{tabthird}{rgb}{1, 1, 0.7}

\newcommand{\model}{Nuvo\xspace}
\newcommand{\normsq}[1]{\left\lVert#1\right\rVert_2^2}

\newcommand{\surfpts}{\mathbf{x}}
\newcommand{\texpts}{\mathbf{u}}

\newcommand{\veryshortarrow}[1][3pt]{\mathrel{%
   \hbox{\rule[\dimexpr\fontdimen22\textfont2-.2pt\relax]{#1}{.4pt}}%
   \mkern-4mu\hbox{\usefont{U}{lasy}{m}{n}\symbol{41}}}}

\newcommand{\differential}{D}

\newcommand{\myparagraph}[1]{ \vspace{3pt}  \noindent {\bf #1}\,\,\,}

\definecolor{cvprblue}{rgb}{0.21,0.49,0.74}
\usepackage[pagebackref,breaklinks,colorlinks,citecolor=cvprblue]{hyperref}
\usepackage{multirow}
\usepackage{tabu}
\usepackage{xcolor,colortbl}
\usepackage{xfrac}

\def\paperID{16583} 
\def\confName{CVPR}
\def\confYear{2024}

\title{\model: Neural UV Mapping for Unruly 3D Representations}

\author{
Pratul P. Srinivasan
\quad
Stephan J. Garbin
\quad
Dor Verbin
\quad
Jonathan T. Barron
\quad
Ben Mildenhall 
\\[5pt]
\centerline{Google Research}
}

\begin{document}

\twocolumn[{%
\renewcommand\twocolumn[1][]{#1}%
\maketitle
\centering
\includegraphics[width=\textwidth]{fig/teaser.pdf}
\captionof{figure}{
        \model is a technique for UV mapping geometry produced by state-of-the-art 3D reconstruction and generation models such as Neural Radiance Fields (NeRFs)~\cite{mildenhall2020nerf}. When applied to such geometry, existing UV mapping algorithms like xatlas~\cite{xatlas} produce fragmented texture atlases (as shown by the chart boundaries marked in orange) that are unusable for tasks like appearance editing. \model produces high-quality editable UV mappings for these 3D models, and is robust to challenging input geometry such as (a) meshes extracted from trained NeRF models and (b) meshes generated by text-to-3D models such as DreamFusion~\cite{poole2022dreamfusion}. \model can even operate directly on (c) NeRF volumetric density fields without requiring a triangulated mesh.\\
        \phantom{}
        }
        \vspace{.15in}
\label{fig:teaser}
}]

\maketitle
\begin{abstract}
Existing UV mapping algorithms are designed to operate on well-behaved meshes, instead of the geometry representations produced by state-of-the-art 3D reconstruction and generation techniques. As such, applying these methods to the volume densities recovered by neural radiance fields and related techniques (or meshes triangulated from such fields) results in texture atlases that are too fragmented to be useful for tasks such as view synthesis or appearance editing. We present a UV mapping method designed to operate on geometry produced by 3D reconstruction and generation techniques. Instead of computing a mapping defined on a mesh's vertices, our method \model uses a neural field to represent a continuous UV mapping, and optimizes it to be a valid and well-behaved mapping for just the set of visible points, \ie only points that affect the scene's appearance. We show that our model is robust to the challenges posed by ill-behaved geometry, and that it produces editable UV mappings that can represent detailed appearance.
\end{abstract}    
\section{Introduction}
\label{sec:intro}

Surface parameterization (``UV mapping'') is the process of flattening a 3D surface onto a plane, and it is a core component of 3D content creation pipelines that enables representing and editing detailed appearance on surface geometry. For complex, real-world meshes, this usually necessitates finding a sequence of cuts such that distortion of the mapping can be minimal. If those cuts result in multiple disconnected components that get packaged into one texture, this is commonly referred to as a ``texture atlas''.

Existing UV mapping algorithms are generally designed to work with well-behaved meshes, such as those created by specialized 3D artists. However, an increasing amount of 3D content does not fall into this category: State-of-the-art methods for reconstructing and generating 3D representations from images or text are based on Neural Radiance Fields (NeRFs), which represent geometry as volumetric fields instead of meshes~\cite{mildenhall2020nerf}. Level sets of volume density are generally not smooth, and triangulating these level sets using techniques such as marching cubes \cite{10.1145/37402.37422} produces meshes with multiple connected components, holes that connect to internal ``hidden'' geometry, and many small ``bumpy'' triangles. More formally, such meshes are typically not manifold or locally smooth, and frequently contain a large number of connected components. Existing UV mapping methods either cannot operate on such meshes or produce heavily fragmented UV atlases that complicate downstream applications. For instance, appearance editing in texture space becomes difficult, and optimization by differentiable rendering is complicated by discontinuities in the surface parameterization which could cause instabilities or necessitate the use of very large textures representations.

We present an approach, which we call \model, that addresses these issues by using neural fields to directly optimize a UV mapping that satisfies the myriad requirements of a well-behaved surface parameterization. Our method simply requires a representation of scene geometry that allows for sampling visible 3D points, and we optimize \model from scratch for each scene by minimizing a set of losses that encourage \model to represent a well-behaved mapping for observed points. Because our method uses point sampling on the surface as its fundamental operation, it can be applied to any implicit surface representation, as well as polygonal meshes, without strong limitations on manifoldness, connectivity, or smoothness. For instance, \model can generate texture atlases directly from NeRF's volume density representation of geometry and it can be applied to extracted meshes while remaining agnostic to the connectivity of the underlying mesh. Because \model's UV mapping representation is not tied to any underlying mesh, it does not suffer from the chart fragmentation issues that can arise when texturing non-smooth meshes with many small triangles.

We test \model on a variety of 3D geometry representations: well-behaved meshes, volume density fields reconstructed by NeRF, meshes extracted from NeRF's volume density fields, and meshes produced by text-to-3D generative models. As illustrated in Figure~\ref{fig:teaser}, \model produces texture atlases that are high-quality with low chart fragmentation, and can therefore be be used to represent and edit detailed surface appearance.
\section{Related Work}
\label{sec:related}

\begin{figure*}[t!]
    \centering
    \includegraphics[width=\linewidth]{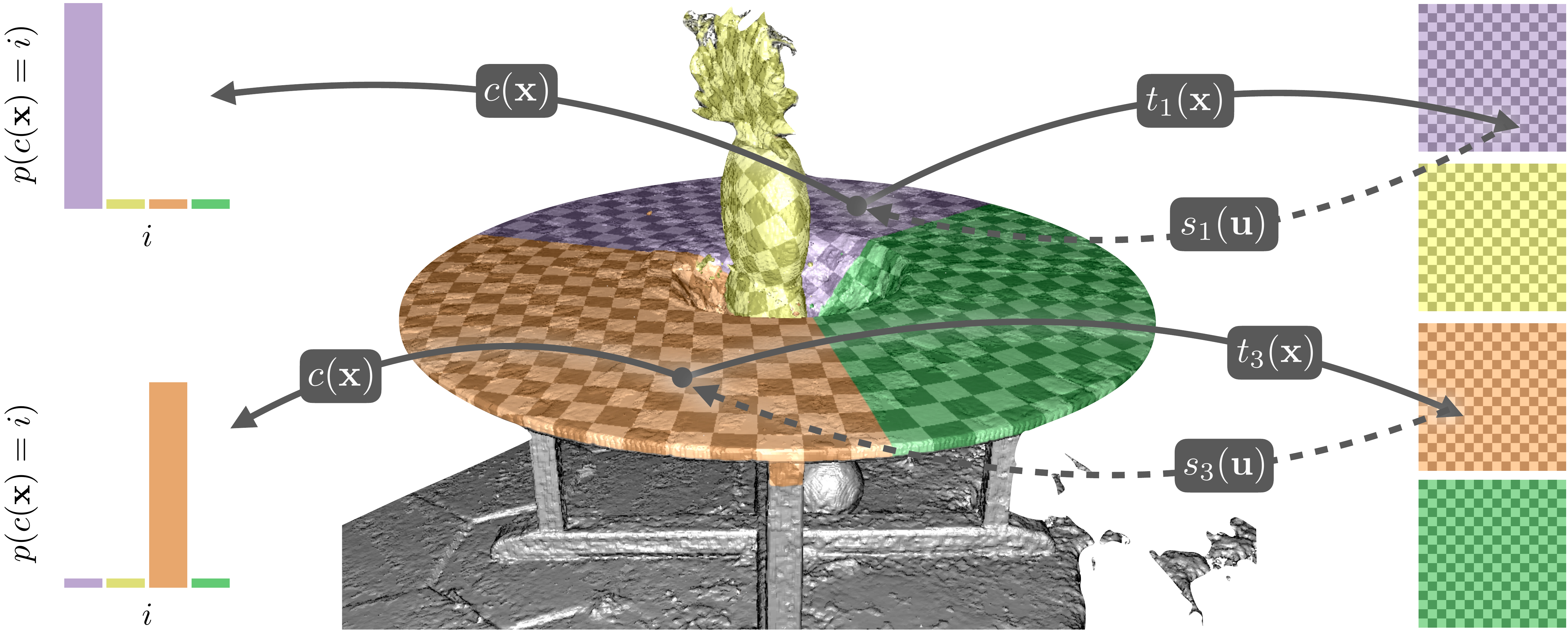}
    \caption{{\bf Overview of \model.} Our model uses a neural field to represent a given scene's UV mapping. Our ``chart assignment'' MLP $c(\cdot)$ outputs probabilities for a categorical distribution over charts for any surface point $\surfpts$, our ``texture coordinate'' MLPs $t_i(\cdot)$ map from 3D points $\surfpts$ to 2D UV coordinates $\texpts$, and our ``surface coordinate'' MLPs $s_i(\cdot)$ map from 2D UV coordinates to 3D points on the surface. Here we visualize \model's learned mappings for charts $i=1$ and $i=3$ in an atlas consisting of $n=4$ charts.}
    \label{fig:method}
\end{figure*}

\myparagraph{Preliminaries}
Surfaces reconstructed by neural field representations can have vastly different topology depending on the scene. As such, triangulated meshes of these representations do not have guarantees on manifoldness, uniformity of triangle areas, smoothness, number of connected components, or genus. Since only developable surfaces (such as a cylinder) allow for isometric mappings to the plane without cuts, our method is designed to optimise both segmentation of the surface, \ie cuts, and the UV mapping, \ie the parameterization, simultaneously.

A mapping is isometric if it preserves both lengths and angles, and conformal if it only preserves angles. Connected components in UV space are referred to as charts, and their collective is called an atlas~\cite{textureAtlasCitation}. Packing charts into an atlas is typically done as a post-processing step, but because \model produces square charts similar to \citet{10.5555/1281957.1281981}, which are trivial to pack, we will not discuss packing in detail. Please refer to~\citet{ShefferBook} and~\citet{10.1007/3-540-26808-1_9} for further review of mesh parameterization.


\myparagraph{Mesh Cutting \& Parameterization}
Mesh parameterization is a long-standing problem \cite{https://doi.org/10.1112/plms/s3-13.1.743, catmull1974subdivision}. However, only a few existing methods address the same problem setting as we do by simultaneously optimizing both cuts and UV mappings.

Many methods instead start with a user-specific boundary and optimize a UV mapping for that specific cut. Least Squares Conformal Mapping (LSCM)~\cite{10.1145/566654.566590} is a widely-used technique that produces conformal maps for a pre-cut mesh. LSCM and related techniques construct and solve a system of equations that is a function of the mesh and its connectivity. This strategy ties these methods to the given mesh topology, and imposes strict requirements like manifoldness. In contrast, \model never explicitly builds and solves such systems and instead relies on sampling points on the surface.

\citet{boundedDistortionSorkine} propose the first method to simultaneously optimize cuts and surface parameterization by starting from from seed triangles and iteratively adding elements until a distortion bound is reached. 

Geometry Images \cite{10.1145/566654.566589} and Multi-Chart Geometry Images \cite{10.5555/882370.882390} encode a 3D triangle mesh on a regularly sampled grid, which allows the geometry to be stored using image compression techniques and facilitates surface re-meshing. Rectangular Multi-Chart Geometry Images improves these algorithms to ensure one-to-one texel assignment across chart boundaries while forming rectangular charts that can be packaged easily \cite{10.5555/1281957.1281981}, like our method. Their cutting and parameterization algorithm is similar to \citet{boundedDistortionSorkine}, but with a different objective function. Unlike \model, these methods require manifold triangle meshes and were developed for well-behaved surfaces with significantly lower complexity than those we address.

AutoCuts~\cite{autoCuts} and OptCuts~\cite{optCuts} generate parameterizations of 3D surfaces that have both minimal cut length and low distortion by alternating the optimization of the mapping distortion (continuous) and cut locus (discrete). While AutoCuts is designed for an interactive workflow, OptCuts operates fully automatically. While OptCuts method can produce mappings with low distortion and simple boundaries on simpler meshes, it is unable to run on high-complexity meshes extracted from neural fields, even after manual cleanup.

A related line of work focuses on improving existing UV parameterizations. \citet{10.1145/2766947} minimize symmetric Dirichlet energy to improve the distortion of an existing parameterization, which can be very efficient \cite{10.1145/2983621} and is more robust than prior work such as \citet{10.1145/566654.566590}. Several recent methods optimize a pre-existing UV mapping to minimise a reconstruction loss using differentiable rendering \cite{jointTexUVOptimisation, https://doi.org/10.1111/cgf.14696}. However, these techniques must be provided with an initial surface parameterization.

\myparagraph{Neural 3D Representations \& Parameterization}
State-of-the-art methods for 3D reconstruction and generation are based on neural fields, which typically represent 3D geometry as some variation of a volumetric density field \cite{Park_2019_CVPR, 10.1145/3592426, yariv2023bakedsdf, barron2023zipnerf} parameterized by some combination of MLPs with positional encoding \cite{tancik2020fourfeat} and feature grids \cite{yu_and_fridovichkeil2021plenoxels,SunSC22, mueller2022instant}. Some recent works focus on finding 2D parameterizations of 3D signals without using explicit triangle meshes. For example, \citet{10.1145/3478513.3480546} convert videos into a set of 2D atlases where an MLP maps from a 3D coordinate in the video to a 2D atlas containing color and transparency. \citet{chen2023uv} and \citet{chen2022AUVNET} learn UV mapping-based models of appearance for the specialized cases of dynamic human models and 3D human faces.
\citet{morreale2021neural} use MLPs to encode single-chart surface mappings for geometry processing problems including establishing correspondences between surfaces.

Our work builds upon NeuTex~\cite{neuTex}, which proposes a joint optimization framework for appearance and surface parameterization. Similar to \model, NeuTex uses a cycle-consistency loss to encourage invertible mappings from 3D to 2D. However, NeuTex and related followups~\cite{neuralgauge,10.1007/978-3-031-19836-6_33} cannot represent atlases with multiple charts, and are therefore not suited to represent mappings for general scenes with multiple connected components.

\section{Method}
\label{sec:method}

Given a representation of scene geometry that allows for sampling visible points in the scene (\eg a NeRF or a mesh), \model generates a UV mapping that partitions the scene geometry into $n$ charts using $2n+1$ MLPs (see Figure~\ref{fig:method}):
\begin{enumerate}
    \item One ``chart assignment'' MLP $c: \mathbb{R}^3 \to \Delta^{n-1}$ assigns points on the surface to charts by mapping from a 3D point to a probability mass function (PMF) of a categorical distribution over the $n$ charts.
    \item A set of $n$ ``texture coordinate'' MLPs $\{ t_i: \mathbb{R}^3 \to [0, 1]^2 \}$, each of which describes chart $i$'s UV mapping from a 3D point to the corresponding 2D ``UV'' texture coordinate (bounded to lie between 0 and 1).
    \item A set of $n$ ``surface coordinate'' MLPs $\{ s_i: [0, 1]^2 \to \mathbb{R}^3 \}$, each of which describes the inverse of each $t_i$ by mapping from a 2D texture coordinate to a 3D point.
\end{enumerate}

\subsection{Losses}
\label{sec:losses}

\begin{figure}[t!]
    \centering
    \includegraphics[width=1.0\linewidth]{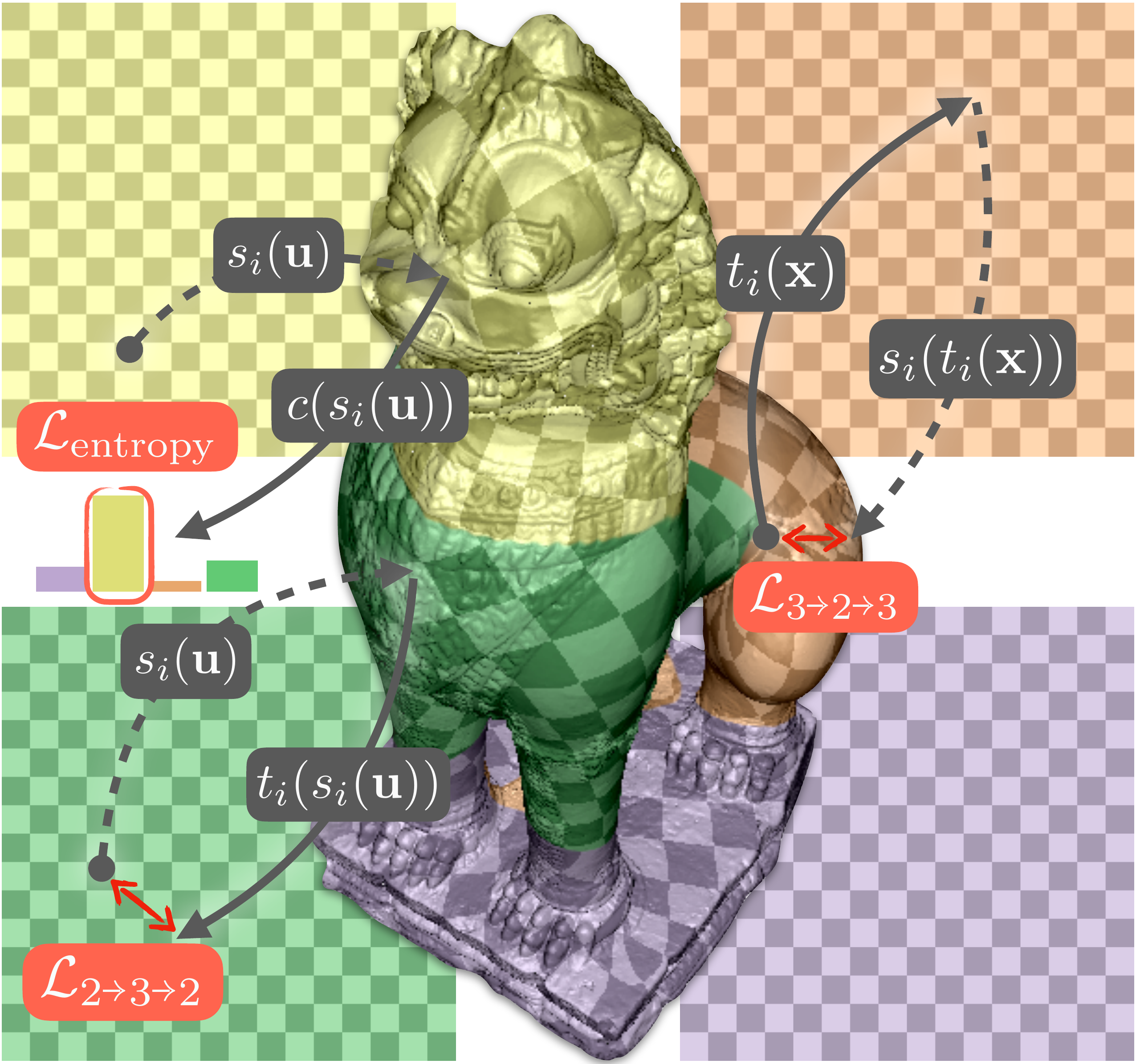}
    \caption{{\bf Visualization of our bijectivity and chart assignment entropy losses.} These losses together encourage \model to partition the scene into a set of bijective chart mappings. $\mathcal{L}_{\mathrm{entropy}}$ maximizes chart $i$'s probability for 3D points $s_i(\texpts)$ that are mapped to from 2D points in chart $i$. $\mathcal{L}_{3\veryshortarrow 2\veryshortarrow 3}$ and $\mathcal{L}_{2\veryshortarrow 3\veryshortarrow 2}$ regularize each chart's mapping to be bijective by encouraging $t_i$ and $s_i$ to be each other's inverse for all 3D surface points $\surfpts$ and all 2D UV points $\texpts$.}
    \label{fig:losses1}
\end{figure}

We optimize \model by minimizing a weighted sum of losses that encourage the recovery of a well-behaved mapping for the scene. The losses are averaged over batches of either random 3D points $\surfpts \in \mathcal{G}$ sampled from the input scene geometry, or random 2D points $\texpts \in \mathcal{T}$ distributed uniformly in texture space $[0, 1]^2$.

\myparagraph{Mapping bijectivity} Each chart's texture coordinate mapping from 3D points on the surface to 2D points on the plane should be approximately bijective, \ie both injective (multiple 3D points should not map to the same texture coordinate) and surjective (the entire texture space should be ``used'' by the mapping). Injectivity is critical, as we require the ability to make separate edits to the appearance of all distinct surface points. Surjectivity is desirable but does not need to strictly hold, as a small amount of unused texture space is tolerable for our purposes.
We encourage bijectivity by minimizing two cycle consistency losses. The first loss minimizes the squared distance travelled by 3D points $\surfpts$ after being mapped to 2D and back:
\begin{equation}
\mathcal{L}_{3\veryshortarrow 2\veryshortarrow 3} = \frac{1}{|\mathcal{G}|} \sum_{\mathbf{x} \in \mathcal{G}} \sum_i c(\surfpts)[i] \cdot \normsq{s_i\lft(t_i\lft(\surfpts\rgt)\rgt) - \surfpts}\,.
\end{equation}
We weight the loss by chart probabilities $c(\cdot)[i]$ (the PMF value corresponding to chart $i$ predicted by the chart assignment MLP $c(\cdot)$) such that the cycle consistency loss for a given point $\surfpts$ under chart $i$ is proportional to the current probability estimate that point $\surfpts$ belongs to chart $i$.
Our second cycle consistency loss minimizes the squared distance travelled by 2D points $\texpts$ after being mapped to 3D and back:
\begin{equation}
\mathcal{L}_{2\veryshortarrow 3\veryshortarrow 2} = \frac{1}{|\mathcal{T}|} \sum_{} \sum_i \normsq{t_i\lft(s_i\lft(\texpts\rgt)\rgt) - \texpts}\,.
\end{equation}
Note that both both cycle consistency losses are necessary to encourage a bijective mapping: $\mathcal{L}_{3\veryshortarrow 2\veryshortarrow 3}$ encourages chart mappings to be injective (one-to-one), but it does not penalize mappings that only map to a subregion of texture space, while $\mathcal{L}_{2\veryshortarrow 3\veryshortarrow 2}$ encourages the entirety of texture space to be covered by an invertible mapping, but it does not penalize mappings that are degenerate for parts of the 3D scene.

\myparagraph{Chart assignment entropy} Intuitively, if we sample a 2D point $\texpts$ in chart $i$, the corresponding 3D point mapped to by surface coordinate MLP $s_i$ should have a chart assignment PMF that is close to a one-hot distribution where the value for chart $i$ is 1 and the rest are 0. We can therefore encourage the chart assignment MLP to confidently partition the 3D scene by minimizing the cross-entropy of the output PMFs with these one-hot distributions:
\begin{equation}
\mathcal{L}_{\mathrm{entropy}} = -\frac{1}{|\mathcal{T}|} \sum_{\mathbf{u} \in \mathcal{T}} \sum_i \log{c(s_i(\texpts))[i]}\,.
\end{equation}

Figure~\ref{fig:losses1} illustrates how our bijective and chart assignment losses together encourage \model to represent a texture atlas that confidently partitions the scene into a set of $n$ bijective charts.

\myparagraph{Surface coordinate} We encourage the surface coordinate MLPs to approximate the input geometry by minimizing the symmetric Chamfer distance between random 2D texture points $\texpts$ mapped to 3D by surface coordinate networks and random 3D points $\surfpts$ sampled from scene geometry:

\begin{align}
\resizebox{0.9\linewidth}{!}{$\displaystyle
\setlength\arraycolsep{1pt}
\begin{matrix} \displaystyle \mathcal{L}_{\mathrm{surface}} & = & \hspace{32pt} \displaystyle \frac{1}{|\mathcal{G}|} & \displaystyle \sum_{\mathbf{x} \in \mathcal{G}} & \!\!\! \underset{\surfpts' \in \bigcup_i s_i(\mathcal{T})}{\min} & \normsq{\surfpts' - \surfpts} &  \\
\displaystyle  & + & \displaystyle \frac{1}{|\bigcup_i s_i(\mathcal{T})|} & \displaystyle \sum_{\surfpts' \in \bigcup_i s_i(\mathcal{T})} & \!\!\! \underset{\surfpts}{\min} & \normsq{\surfpts' - \surfpts} & \,,
\end{matrix}
$}
\end{align}
where $\bigcup_i s_i(\mathcal{T})$ is the union of the 3D points mapped to by all surface coordinate MLPs from random 2D UV points $\texpts \in \mathcal{T}$. This loss is meant to encourage the entire UV domain to be mapped to by points on the 3D geometry. Without ensuring that all 2D texture points map back to the actual 3D surface, it is possible for the MLPs to only represent injective mappings to a subregion of UV space and satisfy $\mathcal{L}_{2\veryshortarrow 3\veryshortarrow 2}$ with injective mappings from 3D regions not on the surface to the remainder of UV space. 

\myparagraph{Chart assignment clustering} We penalize texture atlas fragmentation using a clustering loss that minimizes the distance from each 3D point to the centroids of the points assigned to each chart. We weight the loss by the chart assignment probabilities such that distances from a point $\surfpts$ to the centroid of chart $i$ is penalized proportionally to the probability that point $\surfpts$ belongs to chart $i$:

\begin{equation}
\resizebox{\linewidth}{!}{$\displaystyle
\mathcal{L}_{\mathrm{cluster}} = \frac{1}{|\mathcal{G}|} \sum_{\mathbf{x} \in \mathcal{G}} \sum_i c(\surfpts)[i] \cdot \normsq{\left(
\frac{\sum_{\surfpts' \in \mathcal{G}} c(\surfpts')[i] \cdot \surfpts'  }{\sum_{\surfpts' \in \mathcal{G}} c(\surfpts')[i] }\,
 - \surfpts\right)}\,,
 $}
\end{equation}

\begin{figure}[t]
    \centering
    \includegraphics[width=1.0\linewidth]{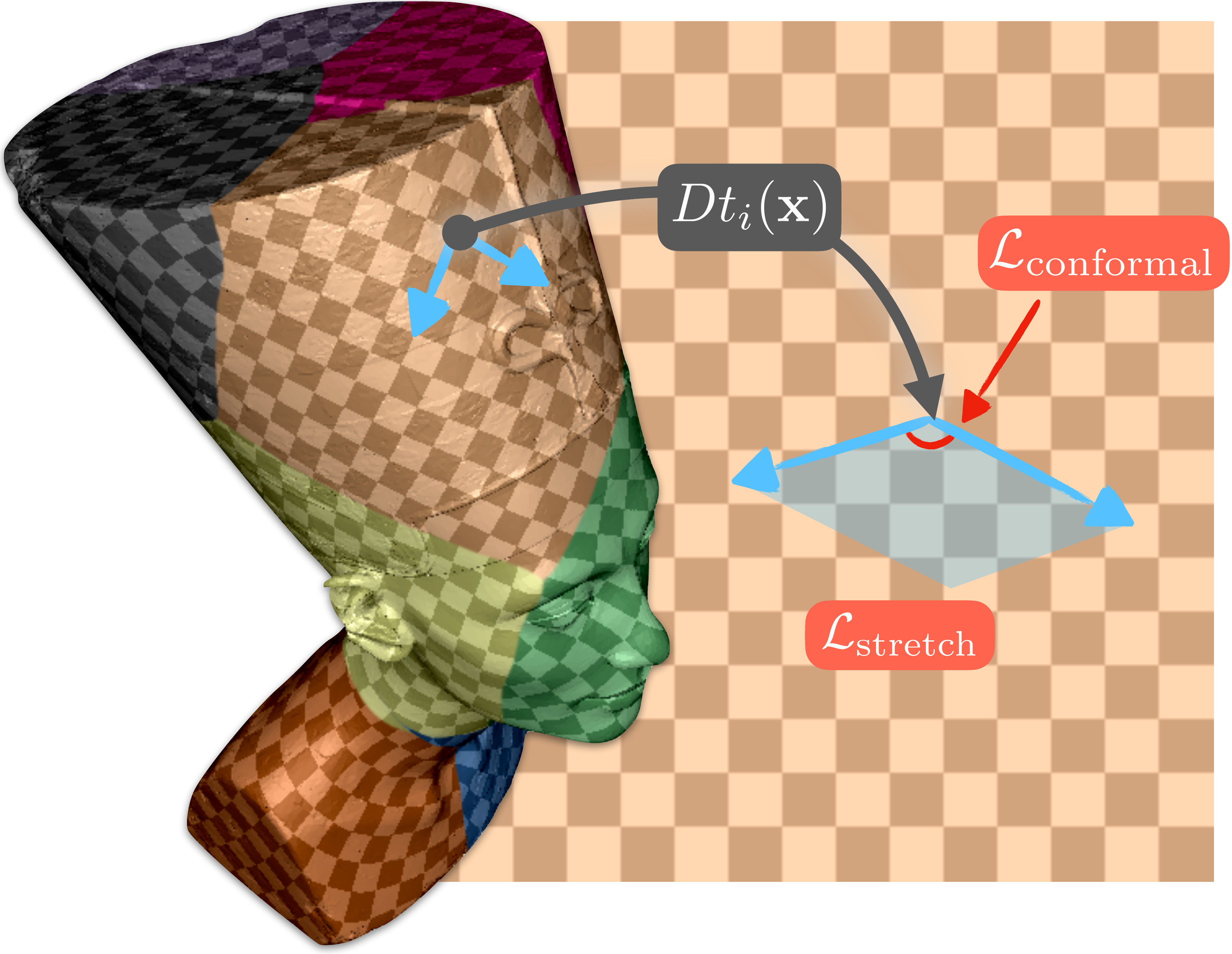}
    \caption{{\bf Visualization of our distortion loss.} The differential of mapping $t_i$, $\differential t_i(\surfpts)$, transforms tangent vectors on the surface to vectors in UV space. $\mathcal{L}_{\mathrm{conformal}}$ encourages the mappings to preserve angles by penalizing the cosine between the differential $\differential t_i(\surfpts)$ applied to orthogonal tangent vectors at $\surfpts$ and $\mathcal{L}_{\mathrm{stretch}}$ encourages the mappings to have uniform stretch by penalizing variation over the scene of the areas of parallelograms spanned by the differentials of orthogonal tangent vectors.}
    \label{fig:losses2}
    \vspace{-0.1in}
\end{figure}

\myparagraph{Distortion} To enable texture editing and to encourage an even allocation of texture resolution over the scene, we want our chart mappings to have low distortion. In particular, we would like the chart mappings to be conformal (angle-preserving), and we would like the amount of stretching or shrinking in the mapping to be uniform, as satisfying these properties ensures that a texture wrapped onto the surface resembles the original texture image.

To this end we impose regularizers on the texture coordinate mappings' differentials, as illustrated in Figure~\ref{fig:losses2}. The differential of mapping $t_i$ at point $\surfpts$, $\differential t_i(\surfpts)$, is a linear mapping of vectors in the 3D surface's tangent space at $\surfpts$ to 2D UV vectors $\texpts$. If a mapping is conformal, applying the differential to orthogonal tangent space vectors should result in orthogonal vectors. If a mapping has uniform stretch, the degree to which $\differential t_i(\surfpts)$ stretches 3D tangent vectors should be constant for all $\surfpts$.

\newcommand{\firstvec}{\epsilon p_{\surfpts}}
\newcommand{\secondvec}{\epsilon q_{\surfpts}}
\newcommand{\difffirstvec}{\differential t_i(\firstvec)}
\newcommand{\diffsecondvec}{\differential t_i(\secondvec)}

We generate random orthogonal tangent space unit vectors, $r_{\surfpts}$ and $v_{\surfpts}$ using the normal at each point $\surfpts$, and then transform the endpoints of $\firstvec$ and $\secondvec$ ($\epsilon=10^{-2}$ in all experiments) to obtain UV vectors $\difffirstvec$ and $\diffsecondvec$ (note that we drop the dependence of the differential on $\surfpts$ for readability). These vectors approximate the mappings' differentials and let us define a distortion loss that is the sum of conformal and uniform stretch regularizers:
$\mathcal{L}_{\mathrm{distortion}}=\mathcal{L}_{\mathrm{conformal}}+\mathcal{L}_{\mathrm{stretch}}$.
Our conformal regularizer is the squared cosine between the transformed vectors, weighted by the probability that $\surfpts$ belongs to chart $i$:
\begin{equation}
\resizebox{\linewidth}{!}{$\displaystyle
\mathcal{L}_{\mathrm{conformal}} = \frac{1}{|\mathcal{G}|} \sum_{\mathbf{x} \in \mathcal{G}} \sum_i c(\surfpts)[i] \cdot \left(\frac{\difffirstvec \cdot \diffsecondvec}{\| \difffirstvec \| \| \diffsecondvec \| }\right)^2\,.
$}
\end{equation}
Our uniform stretch regularizer penalizes the squared difference between the mapping's stretch (the area of the spanned parallelogram) and a single scalar ``average stretch'' scalar parameter $\sigma$ that we optimize for in each scene:
\begin{equation}
\resizebox{\linewidth}{!}{$\displaystyle
\mathcal{L}_{\mathrm{stretch}} \!=\! \frac{1}{|\mathcal{G}|} \sum_{\mathbf{x} \in \mathcal{G}} \sum_i c(\surfpts)[i] \!\cdot\! \Big\lVert \big \lVert \difffirstvec \times \diffsecondvec \big\rVert_2 - \sigma \Big\rVert_2^2 \,.
$}
\end{equation}

This is functionally equivalent to simply minimizing the variance of the stretch over the scene, but we found that alternative to have the undesirable effect of collapsing the texture coordinate MLPs to map all points to the same texture coordinate, while our loss results in stable optimization.

\myparagraph{Texture optimization} We find that optimizing \model's mappings to be usable for representing surface normals helps encourage injectivity. To this end we impose a penalty on the difference between surface normals optimized in UV space and the true surface normals:
\begin{equation}
\mathcal{L}_{\mathrm{texture}} = \frac{1}{|\mathcal{G}|} \sum_{\mathbf{x} \in \mathcal{G}} \sum_i c(\surfpts)[i] \cdot \big \lVert  N_i(t_i(\surfpts)) - n(\surfpts) \big \rVert_2^2 \,,
\end{equation}
where $N_i(\texpts)$ is the value of a pixel grid for chart $i$ at 2D points $\texpts$ (using bilinear interpolation) and $n(\surfpts)$ is the surface normal at $\surfpts$. We jointly optimize pixel grid values in $N_i$ alongside the mapping MLPs to minimize this loss.

\section{Experiments}
\label{sec:experiments}

\begin{figure*}
    \centering
    \includegraphics[width=1.0\linewidth]{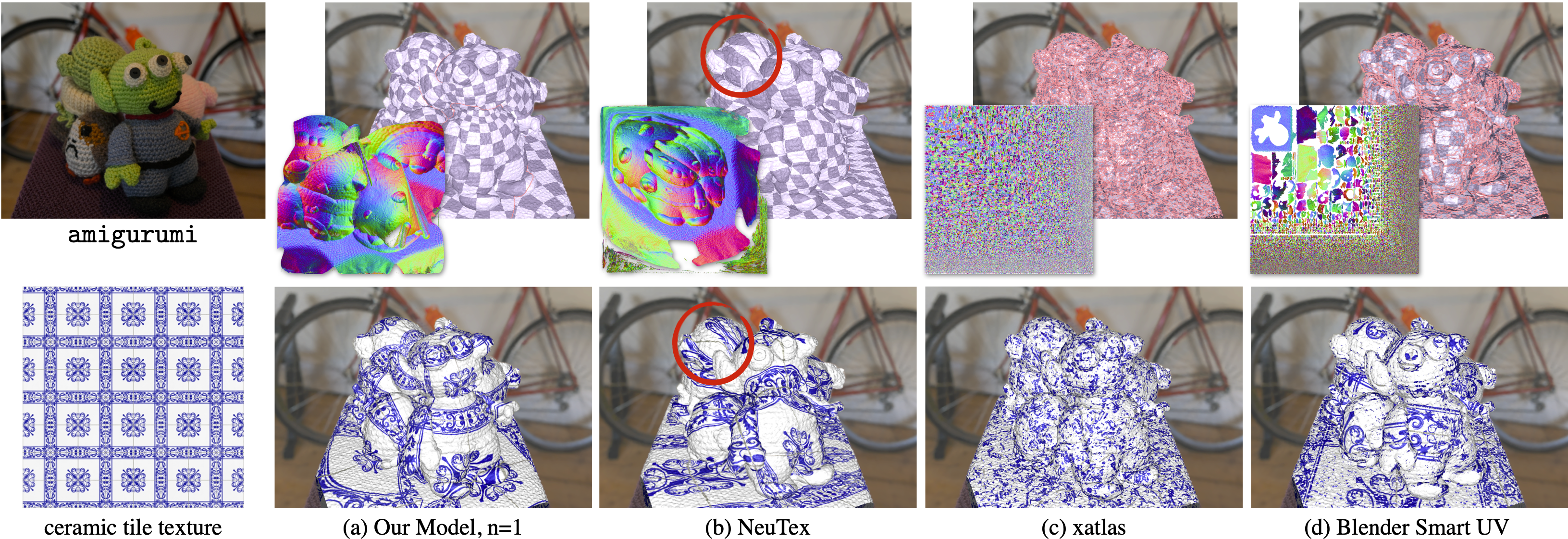}
    \vspace{-0.2in}
    \caption{
    UV mapping and appearance editing comparisons on the \texttt{amigurumi} mesh extracted from Zip-NeRF volume densities.
    (a) \model is able to recover a high-quality editable mapping that supports realistic appearance editing, while (b) NeuTex~\cite{neuTex}'s mappings exhibit significant distortion (red circles), and (c) xatlas'~\cite{xatlas} and (d) Blender's~\cite{blender} atlases are too fragmented for appearance editing. Chart boundaries are marked in orange.}
    \label{fig:comparisons_amigurumi}
\end{figure*}

\begin{table}
\small
\centering
\begin{tabular}{l|cr|cr}
\multicolumn{1}{c}{} & \multicolumn{2}{c}{\texttt{gardenvase}} & \multicolumn{2}{c}{\texttt{amigurumi}} \\
\multicolumn{1}{c|}{} & \multicolumn{1}{c}{PSNR} & \multicolumn{1}{c|}{Params} & \multicolumn{1}{c}{PSNR} & \multicolumn{1}{c}{Params} \\ 
\midrule
BakedSDF~\cite{yariv2023bakedsdf}   &   22.55   &  37.8M  &  28.63  &   59.4M    \\ 
Ours, $n=2$          &   23.11   &  9.1M  &  28.63  &   11.5M   \\ 
Ours, $n=2$ (b)  &   23.09   &  10.1M  &  28.47  &   14.1M   \\ 
Ours, $n=8$          &   23.48   &  9.1M  &  28.81  &   11.5M   \\ 
Ours, $n=8$ (b)  &   22.98   &  10.1M  &  28.42  &   14.1M   \\ 
Ours, $n=32$          &   \cellcolor{tabfirst}23.72   &  9.1M  &  \cellcolor{tabfirst}28.92  &   11.5M   \\ 
Ours, $n=32$ (b)  &  23.41   &   10.1M &  28.67  &   14.1M   \\ 
\end{tabular}
\caption{\textbf{View synthesis comparison.} \model's UV mappings effectively represent detailed surface appearance for view synthesis. We optimize the view-dependent appearance model used in BakedSDF~\cite{yariv2023bakedsdf}, but in UV atlas space instead of on mesh vertices. We evaluate \model's view synthesis performance on two scenes from the Mip-NeRF 360 dataset~\cite{barron2022mip} using $n=2,8,32$ charts, and fix memory usage by using a texture resolution of $256\sqrt{\sfrac{2}{n}}\times256\sqrt{\sfrac{2}{n}}$ for each chart. Additionally, we evaluate ``baked'' versions of \model (indicated with ``b'')  by precomputing and storing the MLP-predicted UV coordinates on mesh vertices to show that \model's MLPs do not need to be kept after optimization, and that our UV maps can be used in standard graphics pipelines.}
\label{table:viewsynth}
\end{table}

\myparagraph{Datasets} We compare UV mappings across four datasets: \texttt{bunny} (72K vertices) and \texttt{lion} (750K vertices) are well-behaved meshes with a single connected component and smooth manifold geometry. \texttt{gardenvase} (1.4M vertices) and \texttt{amigurumi} (2.2M vertices) are meshes extracted by marching cubes from Zip-NeRF~\cite{barron2023zipnerf} reconstructions. These two scenes are captured in the ``Mip-NeRF 360'' style~\cite{barron2022mip} with roughly 185 images per dataset, where one eighth of these are reserved for testing. The meshes of these two scenes have non-smooth geometry, thousands of separate connected components, and non-manifold edges.

When optimizing UV mappings for well-behaved meshes, we sample random 3D points $\surfpts$ uniformly distributed on the surface. When optimizing UV mappings for meshes extracted from NeRF reconstructions, we sample random camera rays that view the mesh, and use their intersection points as $\surfpts$.

\subsection{\model Represents Detailed Appearance}

We first validate \model's ability to produce UV mappings that effectively represent detailed appearance for a given geometry. As our goal is to compute UV mappings for geometry produced by NeRF and other view synthesis approaches, we evaluate the mappings produced by \model and our baselines by measuring how useful they are for view synthesis. 

We conduct this evaluation on the \texttt{gardenvase} and \texttt{amigurumi} datasets. Starting with an optimized NeRF (we use the state-of-the-art Zip-NeRF model~\cite{barron2023zipnerf}), we first extract a mesh using marching cubes~\cite{10.1145/37402.37422}, and then compute a UV mapping using our algorithm. Next, we define a view-dependent appearance model consisting of a diffuse color and three spherical Gaussian view-dependent color lobes (the same appearance model used in BakedSDF~\cite{yariv2023bakedsdf}) on a 2D grid in UV space, and optimize this representation to best reproduce the training images.

Table~\ref{table:viewsynth} compares \model's view synthesis results to those of BakedSDF, which uses the same view-dependent appearance model defined directly on the extracted mesh's vertices. We can see that optimizing our UV mappings for view synthesis performs similarly or even better than optimizing the same appearance model directly on mesh vertices, and we are able to achieve this performance while using less memory than BakedSDF as we do not allocate memory to triangles that do not directly influence view synthesis.

To further demonstrate \model's usefulness in standard graphics pipelines, we ``bake'' the optimized UV coordinates onto the mesh as vertex attributes (we select the chart with the maximum probability for each vertex). This can be thought of as a piecewise linear approximation of the optimized UV mappings. The ``(b)'' rows in Table~\ref{table:viewsynth} demonstrate that even though our UV mappings are optimized as continuous MLP-parameterized functions, baking them onto a mesh incurs a minimal decrease in performance. 

\begin{table*}
\centering
\resizebox{\textwidth}{!}
{
\small
\begin{tabular}{l|cc|cc|cc|cc}
\multicolumn{1}{c}{} & \multicolumn{2}{c}{\texttt{bunny}} & \multicolumn{2}{c}{\texttt{lion}} & \multicolumn{2}{c}{\texttt{gardenvase}} & \multicolumn{2}{c}{\texttt{amigurumi}}\\
\multicolumn{1}{c|}{} & 
\multicolumn{1}{c}{Boundary} & \multicolumn{1}{c|}{Editability} & \multicolumn{1}{c}{Boundary} & \multicolumn{1}{c|}{Editability} & \multicolumn{1}{c}{Boundary} & \multicolumn{1}{c|}{Editability} & \multicolumn{1}{c}{Boundary} & \multicolumn{1}{c}{Editability} \\ 
\midrule
Ours, $n=1$
& 0.987 & \cellcolor{tabsecond}0.946 & 0.988 & \cellcolor{tabsecond}0.914 & 0.986 & 0.682 & 0.983 & \cellcolor{tabfirst}0.836 \\ 
Ours, $n=2$
& 0.982 & 0.937 & 0.977 & 0.890 & 0.974 & \cellcolor{tabfirst}0.859 & 0.983 & \cellcolor{tabsecond}0.813 \\ 
Ours, $n=8$
& 0.970 & 0.923 & 0.964 & 0.882 & 0.933 & \cellcolor{tabsecond}0.780 & 0.966 & 0.809 \\ 
Ours, $n=32$
& 0.924 & 0.860 & 0.903 & 0.811 & 0.823 & 0.704 & 0.927 & 0.784 \\ 
NeuTex~\cite{neuTex}  
& 0.990 & 0.777 & 0.987 & 0.652 & 0.989 & 0.516 & 0.994 & 0.659 \\ 
xatlas~\cite{xatlas}  
& 0.926 & 0.924 & 0.692 & 0.689 & 0.457 & 0.457 & 0.249 & 0.248 \\ 
Blender Smart UV~\cite{blender}    
& 0.876 & 0.837 & 0.680 & 0.657 & 0.778 & 0.764 & 0.603 & 0.581 \\ 
OptCuts~\cite{optCuts}  
& 0.987 & \cellcolor{tabfirst}0.954 & 0.970 & \cellcolor{tabfirst}0.939 & --- & --- & --- & --- \\ 
\end{tabular}
}
\caption{\textbf{Editability comparison.} \model is competitive with the state-of-the-art OptCuts method specifically designed for well-behaved meshes (\texttt{bunny} and \texttt{lion}) and \model significantly outperforms all baselines on the challenging meshes extracted from Zip-NeRF (\texttt{gardenvase} and \texttt{amigurumi}), where OptCuts is unable to compute any UV mapping. The ``Boundary'' metric quantifies the texture atlas fragmentation (higher is better \ie less fragmented), and the ``Editability'' metric quantifies a user's ability to wrap arbitrary new textures on the geometry (higher is better, best method is highlighted in red and second best method is highlighted in orange), which requires that the mapping has both low fragmentation and low distortion.}
\label{table:edit}
\end{table*}

\subsection{\model Produces Editable Mappings}


\begin{figure*}
    \centering
    \includegraphics[width=\linewidth]{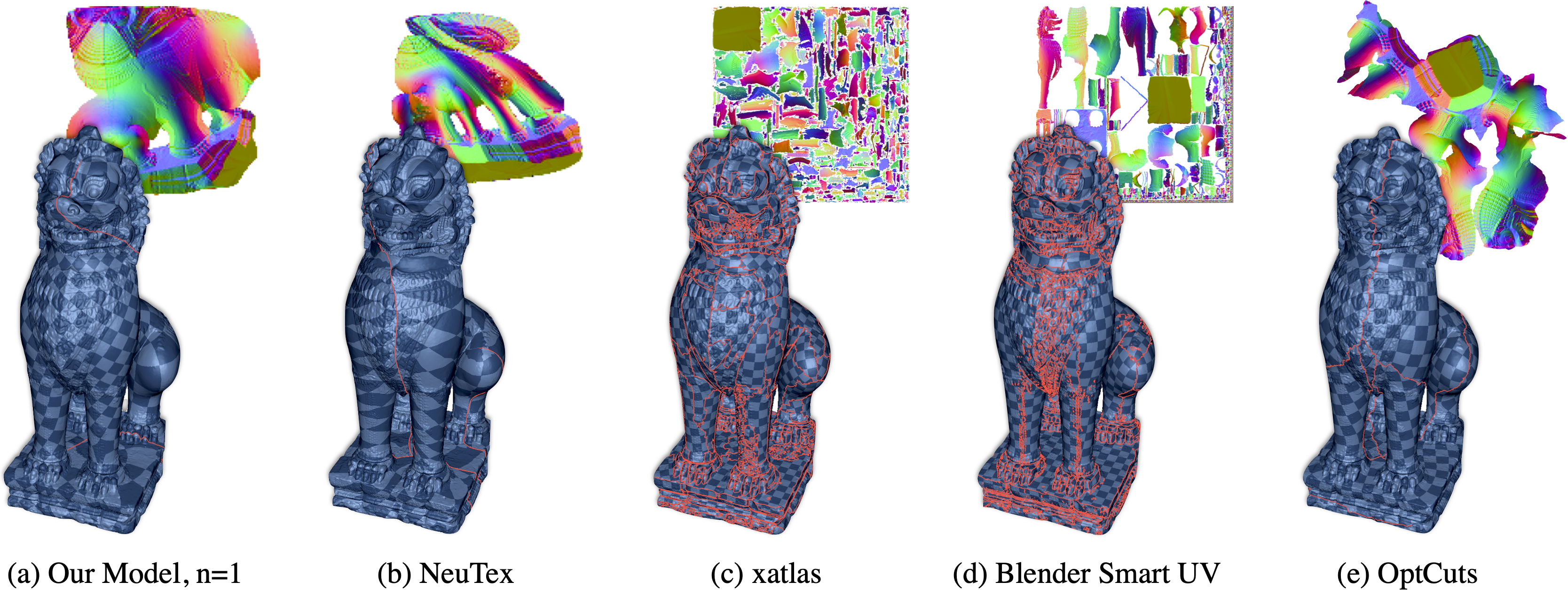}
    \caption{UV mapping comparisons on the \texttt{lion} mesh. (a) \model produces a mapping with a simple boundary and low overall distortion. (b) NeuTex~\cite{neuTex}'s mapping is distorted and therefore challenging to edit. (c) xatlas and (d) Blender produce atlases with significant fragmentation that preclude appearance editing. (e) OptCuts~\cite{optCuts} recovers a high-quality mapping for this well-behaved mesh, but is unable to produce mappings for the more challenging meshes extracted from NeRF.  Chart boundaries are marked in orange.}
    \label{fig:comparisons_lion}
\end{figure*}

\myparagraph{Baselines}
We compare UV mappings produced by our model (using $n$ = 1, 2, 8, and 32 charts) to:
\begin{itemize}
    \item[] \textbf{NeuTex}~\cite{neuTex}, which optimises a single-chart paramterization with a bijective consistency loss ($\mathcal{L}_{3\veryshortarrow 2\veryshortarrow 3}$) and no penalty on distortion; 
    \item[] \textbf{xatlas}~\cite{xatlas}, an open-source library used throughout the 3D content creation industry based on work such as \citet{10.1145/566654.566590} and \citet{10.5555/882370.882390}. Xatlas has been used in many recent reconstruction methods~\cite{Munkberg_2022_CVPR, hasselgren2022nvdiffrecmc};
    \item[] \textbf{Blender Smart UV}~\cite{blender}, a robust industrial tool which we found to behave similarly to ``automatic UV'' methods in proprietary and commercial applications;
    \item[] \textbf{OptCuts}~\cite{optCuts}, the most recent academic method we found for obtaining a UV atlas. OptCuts uses alternating optimization to minimize distortion and cut length.
\end{itemize}
We use the official implementations of xatlas, OptCuts, and Blender 2.93, all with default settings. We were unable to run OptCuts on our more complex scenes as it failed to find a suitable initial UV embedding, even after manually simplifying the meshes by removing non-manifold geometry and multiple connected components.

\myparagraph{Metrics}
We compare UV mappings using two metrics: ``Boundary'' and ``Editability'' (higher is better for both), computed over a test set of randomly-sampled camera viewpoints for each dataset. ``Boundary'' quantifies texture atlas fragmentation by measuring the fraction of rendered pixels that do not lie on chart boundaries. ``Editability'' quantifies a user's ability to wrap arbitrary texture on the geometry. First, we compute the UV coordinates for the vertices of the triangle intersected by each camera ray (using the ``baked'' versions of our MLP-parameterized mappings for fair comparison). Next, we measure each triangle's editability as the average of angular distortion and area distortion metrics of the linear mapping implied by the UV coordinates of the three vertices. The image's total editability score is the average of each pixel's triangle editability masked by chart boundaries;  pixels corresponding to chart boundaries are not considered editable. Please refer to the supplementary materials for a full detailed definition of these metrics. 

\subsection{Ablation Studies}

\begin{figure}
    \centering
    \includegraphics[width=1.0\linewidth]{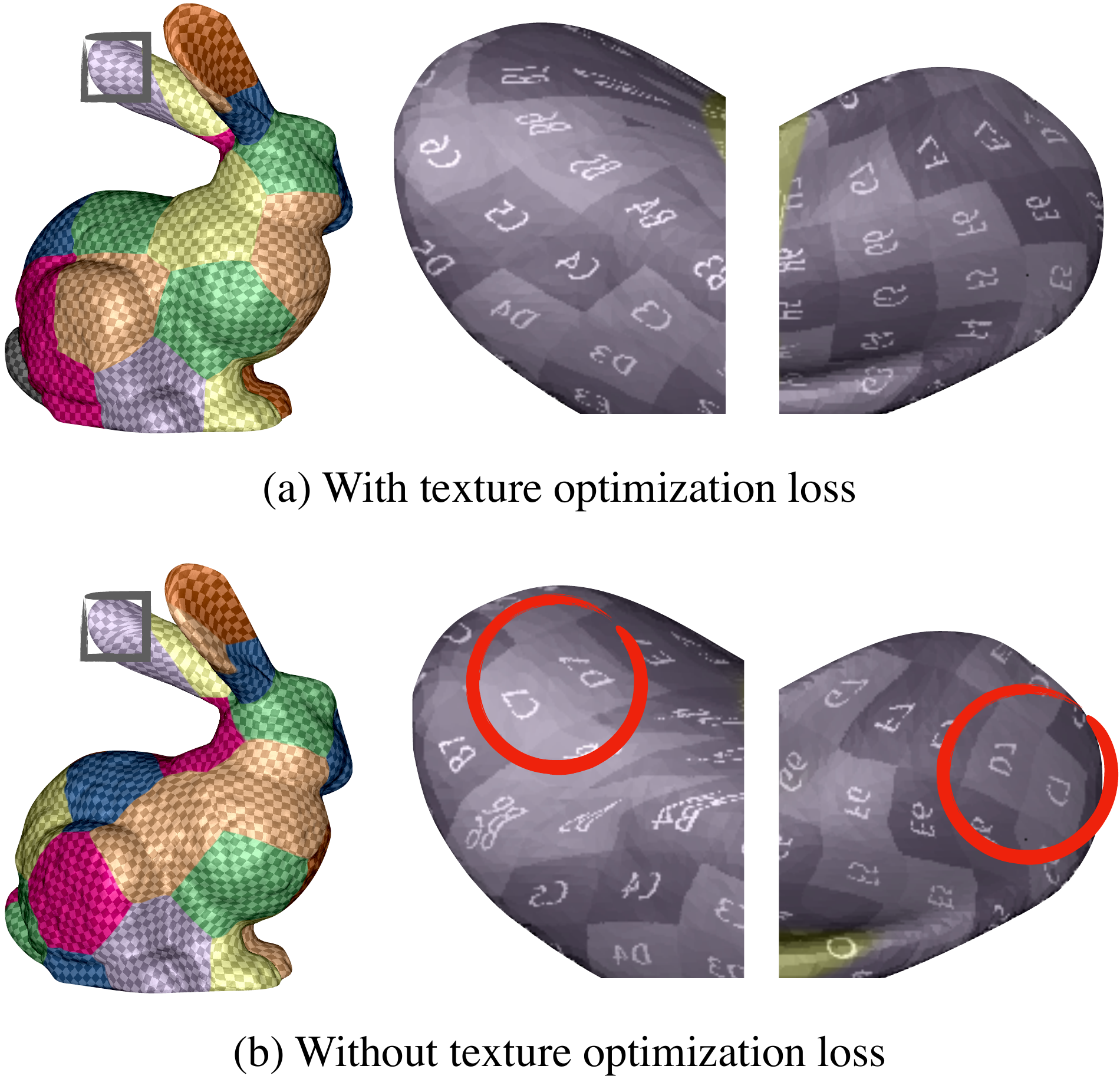}
    \caption{{\bf Texture optimization loss} $\mathcal{L}_{\mathrm{texture}}$ encourages the optimized mappings to be injective (one-to-one) by penalizing UV mappings that cannot accurately represent surface normals. Inspecting the optimized mappings for the bunny's left ear (grey box) from either side, we see that an ablation trained without texture optimization maps both sides of the ear to the same coordinates in UV space; the ``C7'' and ``D7'' UV coordinates (red circles) are visible from both sides of the ear. Our complete model trained with texture optimization does not have this degeneracy.}
    \label{fig:normals}
\end{figure}

The loss ablations in Table~\ref{table:ablation} demonstrate that our full model achieves the best tradeoff of minimizing texture fragmentation (``Boundary''), minimizing distortion (``Stretch'' and ``Conformal''), and utilizing all of UV space (``UV Efficiency''). Figure~\ref{fig:normals} visualizes the effect of our texture optimization loss for encouraging mapping invertibility.

\begin{table}[t]
\resizebox{\linewidth}{!}
{
\begin{tabular}{l|cccc|c}
\multicolumn{1}{c}{} & Bound. & Stretch & Conf. & \multicolumn{1}{c}{UV Eff.} & Avg. \\ 
\midrule
Ours, $n=4$
& 0.976 & 0.976 & 0.940 & 0.846 & \cellcolor{tabfirst}0.935 \\ 
w/o $\mathcal{L}_{3\veryshortarrow 2\veryshortarrow 3}$  
& 0.973 & 0.974 & 0.898 & 0.617 & 0.865 \\ 
w/o $\mathcal{L}_{2\veryshortarrow 3\veryshortarrow 2}$  
& 0.974 & 0.990 & 0.914 & 0.247 & 0.782 \\ 
w/o $\mathcal{L}_{\mathrm{entropy}}$  
& 0.976 & 0.988 & 0.946 & 0.456 & 0.841 \\ 
w/o $\mathcal{L}_{\mathrm{surface}}$ 
& 0.982 & 0.987 & 0.948 & 0.381 & 0.824 \\ 
w/o $\mathcal{L}_{\mathrm{cluster}}$
& 0.953 & 0.979 & 0.947 & 0.785 & 0.916 \\
w/o $\mathcal{L}_{\mathrm{distortion}}$  
& 0.974 & 0.983 & 0.588 & 0.965 & 0.878 \\ 
w/o $\mathcal{L}_{\mathrm{texture}}$
& 0.968 & 0.981 & 0.937 & 0.848 & 0.933 \\ 
\end{tabular}
}
\caption{\textbf{Loss ablations} for \texttt{bunny} using $n=4$. Removing either of the bijectivity, chart assignment entropy, surface coordinate, or chart assignment clustering losses results in mappings that only use a subregion of UV space (low ``UV Efficiency''). Ablating the distortion regularization results in a mapping with much worse conformal distortion. Removing the texture optimization loss does not significantly affect metrics, but can result in non-invertible mappings, as shown in Figure~\ref{fig:normals}.}
\label{table:ablation}
\end{table}

\subsection{Discussion}

Table~\ref{table:edit} demonstrates that \model's UV mappings are competitive with OptCuts' on simpler well-behaved meshes, such as \texttt{lion} shown in Figure~\ref{fig:comparisons_lion}, and significantly better than all baselines for the more challenging cases of geometry extracted from NeRF models, such as \texttt{amigurumi} shown in Figure~\ref{fig:comparisons_amigurumi}. These results are in line with our expectations. OptCuts effectively minimizes both boundary length and distortion, but its optimization procedure is expensive and does not apply to the unstructured non-manifold geometry produced by NeRF. xatlas' and Blender's strategy of starting with many cuts and attempting to merge mappings is not able to produce simple boundaries for non-smooth surfaces, resulting in heavily-fragmented atlases.

\myparagraph{Limitations} 
One of \model's strengths is that it optimizes a UV mapping with point samples instead of explicitly parameterizing the mapping over an entire mesh. However, this makes it harder to absolutely guarantee that it is bijective or that it globally minimizes distortion for a given boundary. Finally, while \model is able to automatically optimize editable UV mappings for challenging geometry, it currently lacks interactive mapping capabilities such as allowing users to specify cut locations or regions for which they would particularly like to minimize distortion. We think that extending \model to address these deficiencies would be fruitful directions for future work.
\section{Conclusion}
\label{sec:conclusion}
We have presented \model, a method that produces editable UV atlases without severe fragmentation and distortion, even for challenging geometry created by 3D reconstruction and generation techniques. By focusing on visible surfaces and parameterizing mappings using neural fields instead of directly on mesh vertices, \model can handle scenes with complexity far beyond the capability of prior approaches. We believe that this work opens up numerous possibilities for creative and artistic editing of reconstructed 3D content.

\section*{Acknowledgements}
We thank Rick Szeliski and Peter Hedman for feedback and comments.

{
    \small
    \bibliographystyle{ieeenat_fullname}
    \bibliography{main}
}

\clearpage
\setcounter{page}{1}
\maketitlesupplementary

\section{Video}

We strongly encourage the reader to watch the supplemental video for additional results.

\begin{figure*}
    \centering
    \includegraphics[width=1.0\linewidth]{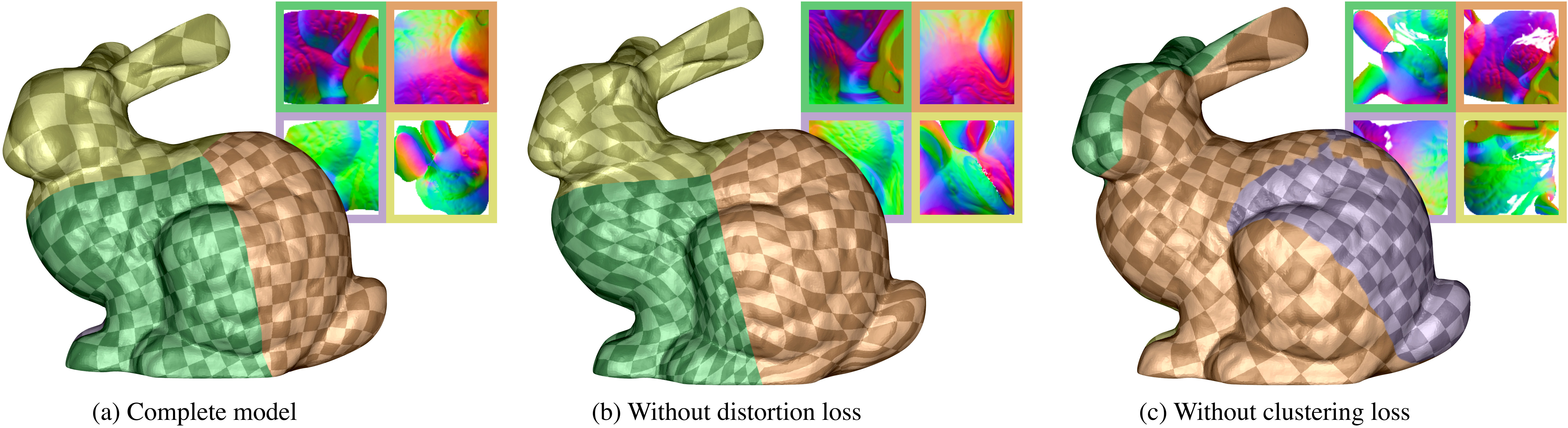}
    \caption{{\bf Loss ablations.} (a) Our complete model recovers a high-quality UV mapping for the bunny model. (b) Removing our distortion loss $\mathcal{L}_{distortion}$ results in a texture atlas with non-uniform warping that hinders editing in 2D texture space. (c) Removing our chart assignment clustering loss $\mathcal{L}_{cluster}$ leads to fragmented and irregularly-shaped charts, which also makes appearance editing cumbersome.}
    \label{fig:ablation}
\end{figure*}

\begin{figure*}
    \centering
    \includegraphics[width=\linewidth]{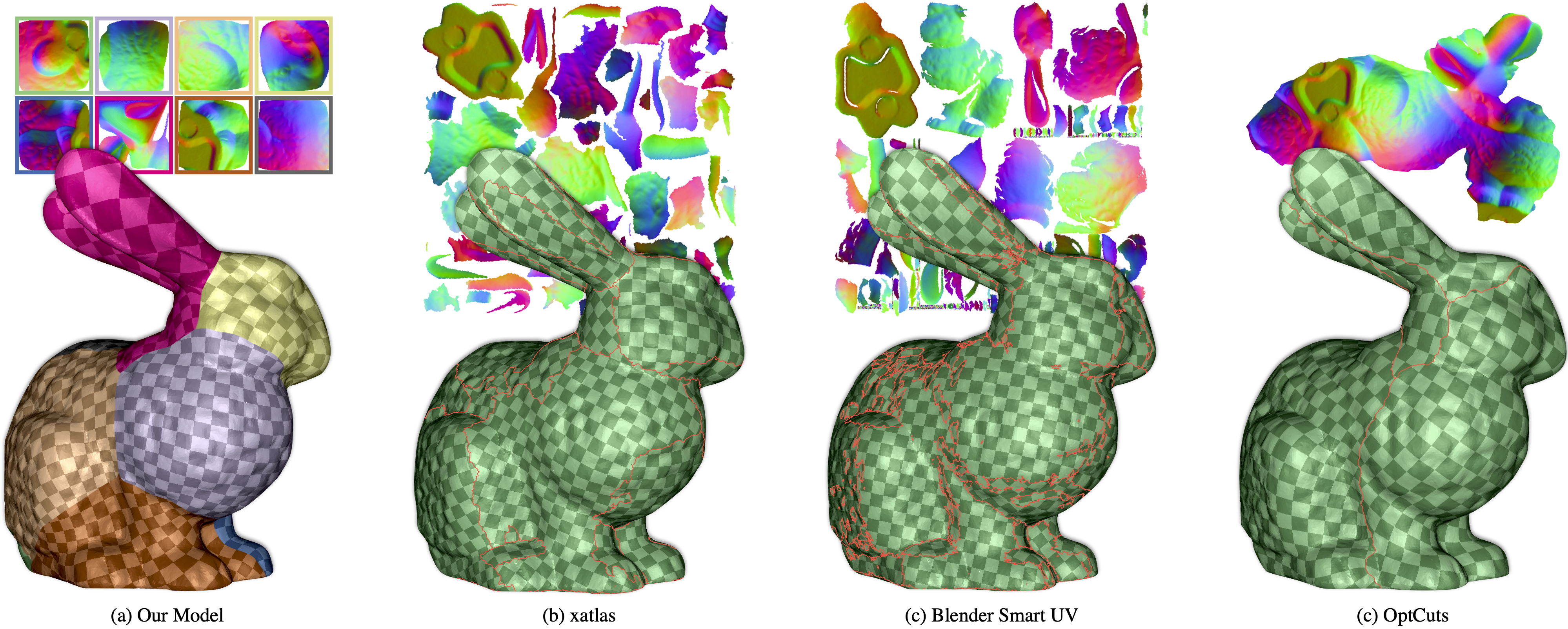}
    \caption{UV mapping comparisons on the \texttt{bunny} mesh. (a) \model produces a mapping with a simple boundary and low overall distortion. (b) NeuTex~\cite{neuTex}'s mapping is distorted and therefore challenging to edit. (c) xatlas and (d) Blender produce atlases with significant fragmentation that preclude appearance editing. (e) OptCuts~\cite{optCuts} recovers a high-quality mapping for this well-behaved mesh, but is unable to produce mappings for the more challenging meshes extracted from NeRF.  Chart boundaries are marked in orange.}
    \label{fig:comparisons_bunny}
\end{figure*}

\begin{table*}
\centering
\resizebox{\textwidth}{!}
{
\small
\begin{tabular}{l|ccc|ccc|ccc|ccc}
\multicolumn{1}{c}{} & \multicolumn{3}{c}{\texttt{bunny}} & \multicolumn{3}{c}{\texttt{lion}} & \multicolumn{3}{c}{\texttt{gardenvase}} & \multicolumn{3}{c}{\texttt{amigurumi}}\\
\multicolumn{1}{c|}{} & 
\multicolumn{1}{c}{Boundary} &
\multicolumn{1}{c}{Stretch} &\multicolumn{1}{c|}{Conf.} & \multicolumn{1}{c}{Boundary} &
\multicolumn{1}{c}{Stretch} &\multicolumn{1}{c|}{Conf.} & \multicolumn{1}{c}{Boundary} &
\multicolumn{1}{c}{Stretch} &\multicolumn{1}{c|}{Conf.} & \multicolumn{1}{c}{Boundary} &
\multicolumn{1}{c}{Stretch} &\multicolumn{1}{c}{Conf.} \\ 
\midrule
Ours, $n=1$ & 0.987 & 0.968 & 0.948 & 0.988 & 0.949 & 0.901 & 0.986 & 0.467 & 0.916 & 0.983 & 0.803 & 0.897 \\ 
Ours, $n=2$ & 0.982 & 0.969 & 0.939 & 0.977 & 0.942 & 0.880 & 0.974 & 0.842 & 0.922 & 0.983 & 0.762 & 0.893 \\ 
Ours, $n=8$ & 0.970 & 0.981 & 0.923 & 0.964 & 0.969 & 0.860 & 0.933 & 0.812 & 0.860 & 0.966 & 0.836 & 0.940 \\ 
Ours, $n=32$ & 0.924 & 0.993 & 0.869 & 0.903 & 0.987 & 0.810 & 0.823 & 0.868 & 0.842 & 0.927 & 0.926 & 0.766 \\ 
NeuTex~\cite{neuTex} & 0.990 & 0.983 & 0.587 & 0.987 & 0.972 & 0.348 & 0.989 & 0.530 & 0.513 & 0.994 & 0.777 & 0.550 \\ 
xatlas~\cite{xatlas} & 0.926 & 0.997 & 0.998 & 0.692 & 0.995 & 0.997 & 0.457 & 0.998 & 0.999 & 0.249 & 0.991 & 0.998 \\ 
Blender Smart UV~\cite{blender} & 0.876 & 0.995 & 0.915 & 0.680 & 0.993 & 0.938 & 0.778 & 0.994 & 0.970 & 0.603 & 0.985 & 0.941 \\ 
OptCuts~\cite{optCuts} & 0.987 & 0.997 & 0.936 & 0.970 & 0.997 & 0.940 & --- & --- & --- & --- & --- & --- \\ 
\end{tabular}
}
\caption{\textbf{Expanded editability comparison with all metrics.} This table contains the ``Boundary'', ``Area'', and ``Angle'' metrics (higher is better for all three) that are used to compute the ``Editability'' metric in Table~\ref{table:edit}.}
\label{table:suppedit}
\end{table*}

\section{Experimental Details}

\myparagraph{MLP Architecture}
We use standard multilayer perceptrons (MLPs) with positional encoding~\cite{mildenhall2020nerf,tancik2020fourfeat} for our chart assignment, texture coordinate, and surface coordinate MLPs. Each MLP has 8 fully-connected layers, each with 256 channels and a ReLU activation function. The chart assignment MLP uses a positional encoding degree of 1, while the texture coordinate and surface coordinate MLPs use a positional encoding degree of 4.

\myparagraph{Point Sampling}
When optimizing UV mappings for well-behaved meshes, we sample random 3D points $\mathbf{x} \in \mathcal{G}$ uniformly distributed on the surface by first sampling random triangles uniformly in area using inverse transform sampling, and then sampling random points on those triangles.

When optimizing UV mappings for meshes extracted from NeRF volume density fields (\eg Zip-NeRF or DreamFusion), we first sample random camera rays (from training images for Zip-NeRF and from random images on a circle for DreamFusion) and then use the intersection points of these rays with the mesh.

When optimizing UV mappings directly on NeRF volume density fields, we first sample random camera rays from the training images, evaluate the Zip-NeRF model to compute a set of sampling points along each ray along with corresponding volume rendering weights, and finally resample a set of ``surface'' points using the normalized total volume rendering weight distribution over all rays. 

\myparagraph{Optimization}
We optimize \model for each scene by minimizing a weighted sum of the losses described in the main paper, using the following weights: $1$ for $\mathcal{L}_{3\veryshortarrow 2\veryshortarrow 3}$, $1$ for $\mathcal{L}_{2\veryshortarrow 3\veryshortarrow 2}$, $0.04$ for $\mathcal{L}_{\mathrm{entropy}}$, $10$ for $\mathcal{L}_{\mathrm{surface}}$, $0.5$ for $\mathcal{L}_{\mathrm{cluster}}$, $0.4$ for $\mathcal{L}_{\mathrm{conformal}}$, and $0.1$ for $\mathcal{L}_{\mathrm{stretch}}$.
We optimize all parameters using the Adam optimizer with a cosine decay schedule, with starting learning rates of $10^{-4}$ for MLP parameters, $0.1$ for the scalar value $\sigma$, and $0.04$ for texture grids $\{ N_i(\cdot) \}$. 

\myparagraph{Runtimes}
Optimizing \model takes $\sim$20 minutes for running on meshes, and $\sim$40 for directly running on NeRF volume density fields. OptCuts took over 2 hours to compute a mapping for the \texttt{bunny} mesh (72K vertices) and over 35 hours for the \texttt{lion} mesh (750K vertices). Since the NeRF scenes are significantly more complex (1.4 million vertices for \texttt{gardenvase} and 2.2 million vertices for \texttt{amigurumi}), it is reasonable to believe that OptCuts would not have been able to compute a result in a reasonable time, even if the unruly nature of the NeRF geometry did not cause the system to fail. The xatlas package took $\sim$2 minutes to run on the \texttt{bunny} mesh, and $\sim$40 minutes to run on the \texttt{lion} mesh.

\section{Metrics}

As mentioned in the main paper, the ``Boundary'' metric quantifies texture atlas fragmentation by measuring the fraction of rendered pixels that do not lie on chart boundaries. We consider any pixel in a rendering as a boundary pixel if its UV coordinate is further than $0.1$ from that of any adjacent pixel, and if it does not lie on a depth discontinuity.

The ``Stretch'' metric for each rendered image quantifies the area distortion for the triangles intersected by each pixel's ray. We first compute the stretch of each triangle as the 2D area of the triangle in UV space divided by the 3D area of the triangle on the mesh. The stretch metric is defined as 1 minus the median of the differences between each triangle's stretch and the median triangle stretch. Thus, UV mappings that have uniform stretch over the entire scene will have a score of 1.

The ``Conformal'' metric for each rendered image quantifies the angular distortion for the triangles intersected by each pixel's ray. We compute tangent and bitangent vectors for each triangle, which are simply the two vectors lying on each triangle that correspond to the positive U and V directions. The conformal metric is defined as 1 minus the median cosine between these tangent vectors. Thus, UV mappings that are conformal and do not introduce angular distortion will have a score of 1.

As discussed in the main paper, the ``Editability'' metric reported in Table~\ref{table:edit} is the mean of the ``Stretch'' and ``Conformal'' metrics, each weighted by the ``Boundary'' metri: $\text{Editability} = \text{Boundary} \cdot \frac{1}{2}(\text{Stretch} + \text{Conformal})$. Thus, ``editable'' UV mappings should have both low atlas fragmentation (a boundary metric close to 1) and low distortion (stretch and conformal metrics close to 1).

\section{Synthetic Data Sources}

The \texttt{lion} mesh (downloaded from \url{threedscans.com}) is a mesh created from a 3D scan of the ``Bayon Lion'' statue in the Mus\'ee Guimet in Paris, France, taken from the Preah Khan Kompong Svay temple complex in Preah Vihear, Cambodia. The \texttt{bunny} mesh is the ``Stanford Bunny'' mesh created from a 3D scan by the Stanford University Computer Graphics Laboratory, downloaded from \url{github.com/alecjacobson/common-3d-test-models}. The statue mesh shown in Figure~\ref{fig:losses2} of the main paper is the ``Bust of Queen Nefertiti'' mesh (downloaded from \url{github.com/alecjacobson/common-3d-test-models}) created from a 3D scan in the Neues Museum in Berlin of the statue created by Thutmose. Our supplementary video additionally contains the ``Hydria Apothecary Vase'' mesh (downloaded from \url{sketchfab.com/3d-models/hydria-apothecary-vase-7d6938c0c0b54b06a0210a982a73023e}) from the Pharmacy Museum in the Jagiellonian University Medical College of Krak\'ow, Poland, digitized by the Regional Digitalisation Lab of the Malopolska Institute of Culture in Krak\'ow, Poland. The tile texture used in Figure~\ref{fig:comparisons_amigurumi} of the main paper was downloaded from \url{https://ambientcg.com/view?id=Tiles101}.

\section{Ablation Visualization}

Figure~\ref{fig:ablation} visualizes the effects of ablating our distortion or clustering losses. Table~\ref{table:ablation} contains quantitative results for all loss ablations.

\section{Additional Qualitative Results}

Figure~\ref{fig:comparisons_bunny} qualitatively compares UV mappings from our method to baselines, and demonstrates that \model produces UV mappings that are better-suited for editing than those produced by existing methods.

\section{Expanded Editability Comparison}

Table~\ref{table:suppedit} contains the full set of metrics used to calculate the ``Boundary'' and ``Editability'' metrics for Table~\ref{table:edit} in the main paper.

\end{document}